# Analysis of Three-Dimensional Protein Images

**Laurence Leherte** LAURENCE.LEHERTE@SCF.FUNDP.AC.BE
*Laboratoire de Physico-Chimie Informatique*
*Facultés Universitaires Notre-Dame de la Paix*
*Namur, Belgium*

**Janice Glasgow** JANICE@QUCIS.QUEENSU.CA
**Kim Baxter** BAXTER@QUCIS.QUEENSU.CA
*Department of Computing and Information Science*
*Queen's University, Kingston, Ontario, Canada, K7L 3N6*

**Evan Steeg** STEEG@QUCIS.QUEENSU.CA
*Molecular Mining Corp., PARTEQ Innovations*
*Queen's University , Kingston, Ontario, Canada, K7L 3N6*

**Suzanne Fortier** FORTIERS@QUCDN.QUEENSU.CA
*Departments of Computing and Information Science and Chemistry*
*Queen's University, Kingston, Ontario, Canada, K7L 3N6*

## Abstract

A fundamental goal of research in molecular biology is to understand protein structure. Protein crystallography is currently the most successful method for determining the three-dimensional (3D) conformation of a protein, yet it remains labor intensive and relies on an expert's ability to derive and evaluate a protein scene model. In this paper, the problem of protein structure determination is formulated as an exercise in *scene analysis.* A computational methodology is presented in which a 3D image of a protein is segmented into a graph of critical points. Bayesian and certainty factor approaches are described and used to analyze critical point graphs and identify meaningful substructures, such as $\alpha$-helices and $\beta$-sheets. Results of applying the methodologies to protein images at low and medium resolution are reported. The research is related to approaches to representation, segmentation and classification in vision, as well as to top-down approaches to protein structure prediction.

## 1. Introduction

Crystallography plays a major role in current efforts to characterize and understand molecular structures and molecular recognition processes. The information derived from crystallographic studies provides a precise and detailed depiction of a molecular scene, an essential starting point for unraveling the complex rules of structural organization and molecular interactions in biological systems. However, only a small fraction of the currently known proteins have been fully characterized.

The determination of molecular structures from X-ray diffraction data belongs to the general class of image reconstruction exercises from incomplete and/or noisy data. Research in artificial intelligence and machine vision has long been concerned with such problems. Similar to the concept of visual scene analysis, *molecular scene analysis* is concerned with the processes of reconstruction, classification and understanding of complex images. Such analyses rely on the ability to segment a representation of a molecule into its meaningful parts, and on the availability of *a priori* information, in the form of rules or structural templates, for interpreting the partitioned image.

A crystal consists of a regular (periodic) 3D arrangement of identical building blocks, termed the unit cell. A crystal structure is defined by the disposition of atoms and molecules within this fundamental repeating unit. A given structure can be solved by interpreting an electron density image of its unit cell content, generated – using a Fourier transform – from the amplitudes and phases of experimentally derived diffraction data. An electron density map is a 3D array of real values that estimate the electron density at given locations in the unit cell; this information gives access to the structure of a protein[1]. Unfortunately, only the diffraction amplitudes can be measured directly from a crystallographic experiment; the necessary phase information for constructing the electron density image must be obtained by other means[2]. This is the classic *phase problem* of crystallography.

In contrast to small molecules (up to 150 or so independent, non-hydrogen atoms), the determination of protein structures (which often contain in excess of 3000 atoms) remains a complex task hindered by the phase problem. The initial electron density images obtained from crystallographic data for these macromolecules are typically incomplete and noisy. The interpretation of a protein image generally involves mental pattern recognition procedures where the image is segmented into features, which are then compared with anticipated structural motifs. Once a feature is identified, this partial structure information can be used to improve the phase estimates resulting in a refined (and eventually higher-resolution) image of the molecule. Despite recent advances in tools for molecular graphics and modeling, this iterative approach to image reconstruction is still a time consuming process requiring substantial expert intervention. In particular, it depends on an individual's recall of existing structural patterns and on their ability to recognize the presence of these motifs in a noisy and complex 3D image representation.

The goal of the research described in this paper is to facilitate the image reconstruction processes for protein crystals. Towards this goal, techniques from artificial intelligence, machine vision and crystallography are integrated in a computational approach for the interpretation of protein images. Crucial to this process is the ability to locate and identify meaningful features of a protein structure at multiple levels of resolution. This requires a simplified representation of a protein structure, one that preserves relevant shape, connectivity and distance information. In the proposed approach, molecular scenes are represented as 3D annotated graphs, which correspond to a trace of the main and side chains of the protein structure. The methodology has been applied to ideal and experimental electron density maps at medium resolution. For such images, the nodes of the graph correspond to amino acid residues and the edges correspond to bond interactions. Initial experiments using low-resolution electron density maps demonstrate that the image can be segmented into protein and solvent regions. At medium resolution the protein region can be further segmented into main and side chains and into individual residues along the main chain.

1. Strictly speaking, the diffraction experiment provides information on the ensemble average over all of the unit cells.
2. Current solutions to the phase problem for macromolecules rely on gathering extensive experimental data and on considerable input from experts during the image interpretation process.

Furthermore, the derived graph representation of the protein can be analyzed to determine secondary structure motifs within the protein.

The paper presents a brief overview of protein structure and the problem of analyzing a molecular scene. The processes of protein segmentation and secondary structure identification are described, along with experimental results. Related and ongoing research issues are also presented.

## 2. Protein Structure

A fundamental goal of research in molecular biology is to understand protein structure and function. In particular, structure information is essential for medicine and drug design. In this section we review some of the concepts involved in protein structure. These concepts will be used later in describing our computational approach to protein structure determination.

A protein is often characterized as a linear list of *amino acids* called the *primary structure*, or sequence, for the protein. All of the naturally occurring amino acids have a similar backbone structure, consisting of a central carbon atom ($C_\alpha$), an amino group ($NH_2$) and a carboxyl group ($COOH$). They are distinguished from one another by their varying side chains that are connected to the $C_\alpha$ atom. Figure 1 illustrates alternative representations for the amino acid *alanine*, where Figure 1(c) displays its side chain consisting of a carbon and three hydrogen atoms. Associated with the side chain of an amino acid are properties such as hydrophobicity (dislikes water), polarity, size and charge.

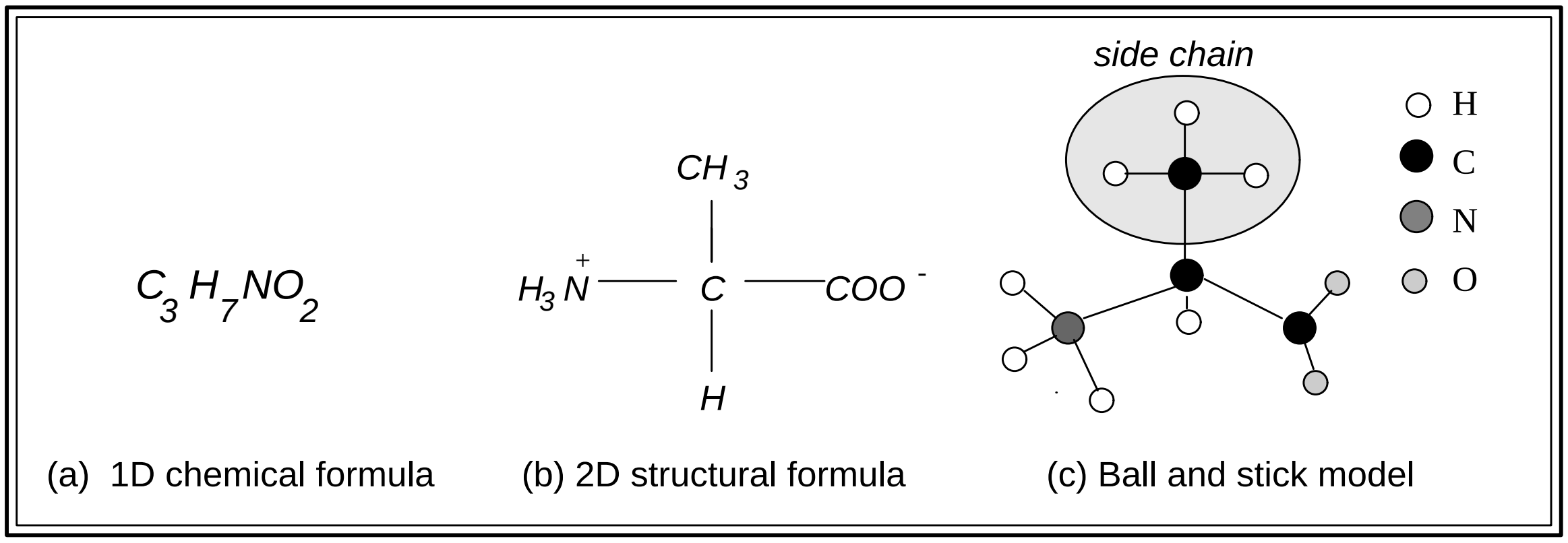

Figure 1: Representations of the amino acid *alanine*.

Adjacent amino acids in the primary structure for a protein are linked together by peptide bonds to form a polypeptide main chain, or backbone, from which the various side chains project (see Figure 2). The carboxyl group of one amino acid joins with the amino group of another to eliminate a water molecule ($H_2O$) and form the peptide bond.

A *secondary structure* of a protein refers to a local arrangement of a polypeptide subchain that takes on a regular 3D conformation. There are two commonly recurring classes of secondary structure: the $\alpha$*-helix* and the $\beta$*-sheet.* An $\alpha$-helix occurs as a corkscrew-shaped conformation, where amino acids are packed tightly together; $\beta$-sheets consist of linear strands (termed $\beta$-strands) of amino acids running parallel or antiparallel to one another (see Figure 3). These secondary structure motifs are generally linked together by less regular structures, termed loops or turns. As will be discussed later in the paper, the characterization of a subsequence of amino acids as an $\alpha$-helix or $\beta$-sheet can be determined by a geometric analysis of the distance and angle relations among local subsequences of amino acids.

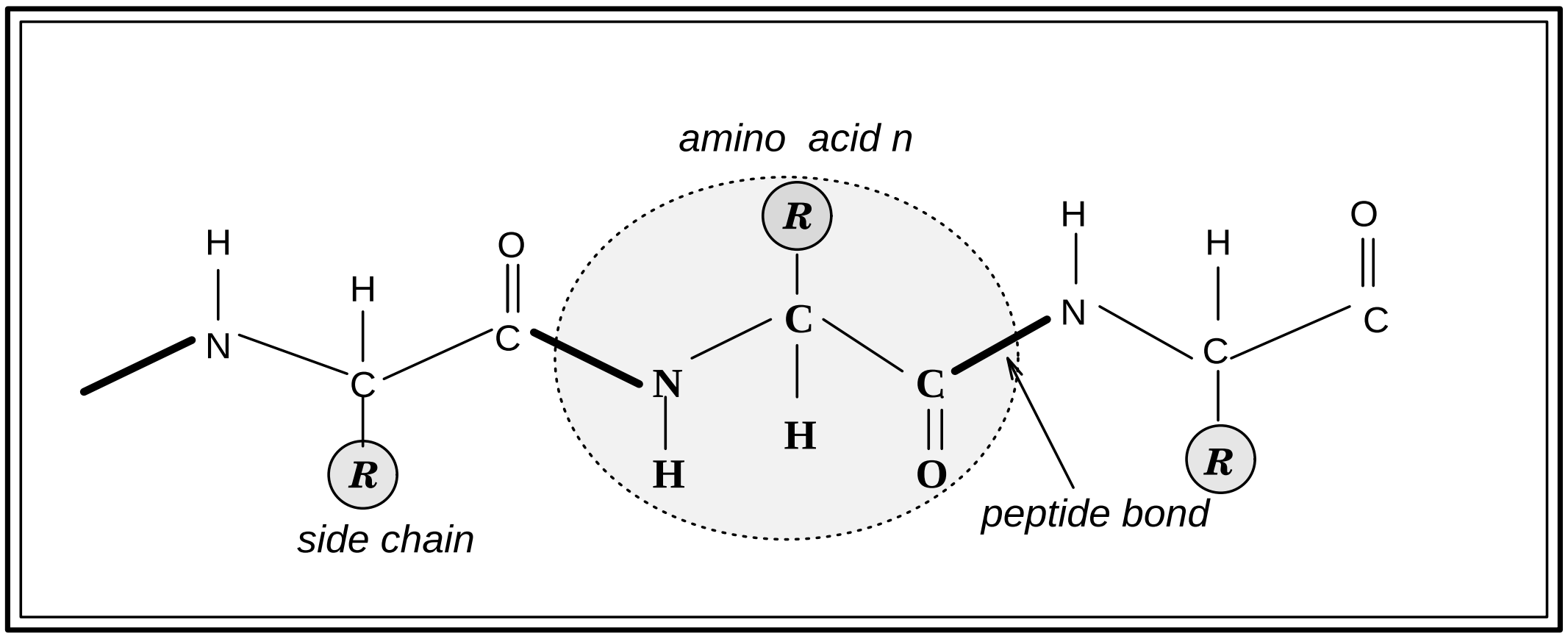

Figure 2: Proteins are built by joining together amino acids using peptide bonds.

The global conformation of a protein is referred to as its *tertiary structure*. The way in which proteins adopt a particular folding pattern depends upon the intramolecular interactions that occur among the various amino acid residues, as well as upon the interaction of the molecule with solvent (water). Two types of intramolecular interactions are often referred to in order to describe the secondary or tertiary structure of a protein. The first type is a *hydrogen bond*, which results from the sharing of a hydrogen atom between residues. $\alpha$-helices and $\beta$-sheets can both be described in terms of regular and specific hydrogen bond networks. Figure 3 illustrates a portion of a $\beta$-sheet in which hydrogen bond interactions link together parallel $\beta$-strands. Additional stability to the 3D structure of a protein is provided by *disulfide bridges*. This second type of interaction is a result of a chemical bond occurring between two sulfur atoms carried by cysteine amino acid residues. These bonds are energetically stronger than hydrogen bonds and contribute to stability under extreme conditions (temperature, acidity, etc.).

Molecular biology is concerned with understanding the biological processes of macromolecules in terms of their chemical structure and physical properties. Crucial to achieving this goal is the ability to determine how a protein folds into a 3D structure given its known sequence of amino acids. Despite recent efforts to predict the 3D structure of a protein from its sequence, the folding problem remains a fundamental challenge for modern science. Since the 3D structure of a protein cannot as yet be predicted from sequence information alone, the experimental techniques of X-ray crystallography and nuclear magnetic resonance currently provide the only realistic routes for structure determination. In the remainder of the paper we focus on a computational approach for the analysis of protein images generated from crystallographic data.

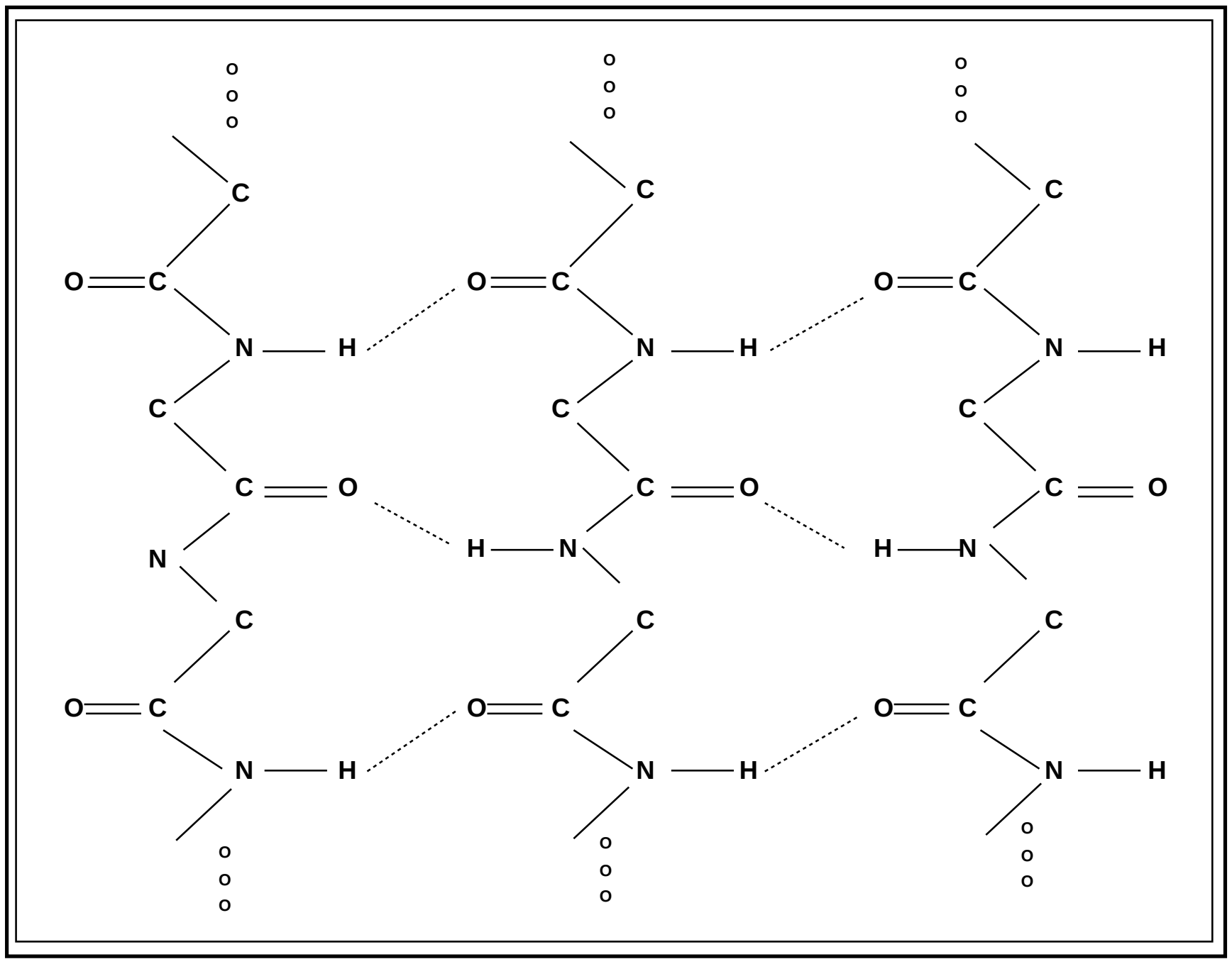

Figure 3: Hydrogen bonds (dotted lines) link three individual $\beta$-strands (linear sequences of amino acids in the main chain of the protein) to form a parallel $\beta$-sheet.

## 3. Scene Analysis

Research in machine vision has long been concerned with the problems involved in automatic image interpretation. Marr (1982) defines computational vision as "the process of discovering what is present in the world, and where it is". Similar to visual scene analysis, molecular scene analysis is concerned with the processes of reconstruction, classification and understanding of complex images. This section presents the problem of molecular scene analysis in the context of related research in machine vision and medical imaging.

Early vision systems generally include a set of processes that determine physical properties of 3D surfaces from 2D arrays. These input arrays contain pixel values that denote properties such as light intensity. Considerable research has been carried out on extracting 3D information from one or more 2D images. A review of the application of stereo vision to sets of 2D images can be found in (Faugeras, 1993). The principles of stereo vision have also been applied to moving images (Zhang & Faugeras, 1992). Range data provides explicit depth information about visible surfaces in the form of a 2D array (depth map). Depending on the application, both surface or/and volume fitting techniques are applied to these images. A more complete review of vision techniques can be found elsewhere (e.g., (Arman & Aggarwal, 1993; Besl & Jain, 1986; Jain & Flynn, 1993)).

Similar to the crystallography problem, medical imaging techniques require genuine 3D data: magnetic resonance imaging (MRI) provides detailed high-resolution information about tissue density; emission computed tomography (ECT), which includes positron emission tomography (PET) and single photon emission computed tomography (SPECT), gives noisy, low-resolution information about metabolic activity. X-ray computed tomography (CT) and ultrasound also provide 3D density data. These methods can be used to obtain a series of 2D images (slices) which, when properly aligned, provide a 3D grid. Because the interslice spacing may be much larger than the spacing between grid points in each slice, alignment of the slices followed by interpolation between the slices is one area of research.

The low-level segmentation of medical images typically uses 3D extensions of 2D techniques. Edge detection becomes surface detection, region growing defines volumes instead of areas. Many applications typically do not require detailed, high-level models. Surface information can be used to construct models for simulations, volumes and surfaces provide structural measurements. Higher-level models are used in the construction of "templates" for pattern matching. One interesting aspect of medical imaging is the availability of *a priori* knowledge, either from a database of similar structures, or from images of the same region in different modalities (e.g. MRI images of a brain can be used to guide segmentation of a lower-resolution PET image). One modality may also provide information that is not clear in another modality. There is considerable research on "registration" of images – aligning or overlaying two 3D images to combine information. Segmentation and identification of matching "landmarks" is important for such image representations.

Unlike input for the vision and medical imaging problems, the crystallographic experiment yields data that define a 3D function, which allows for the construction of a 3D array of voxels of arbitrary size[3]. An interpretation of the 3D atomic arrangement in a crystal structure is readily available for small molecules using the data generated from X-ray

3. Comparatively, machine vision techniques generally provide 2D image representations and range data only provide surface information. Medical imaging techniques may result in a 3D grid, but the spacing diffraction techniques. Given the magnitudes of the diffracted waves and prior knowledge about the physical behavior of electron density distributions, probability theory can be applied to retrieve phase information. Once magnitudes and phases are known, the spatial arrangement of atoms within the crystal can be obtained by a Fourier transform. The electron density function that is obtained, $\rho(x, y, z)$, is a scalar field visualized as a 3D grid of real values called the *electron density map*. High-density points in this image are associated with the atoms in the small molecule.

The construction of an interpretable 3D image for a protein structure from diffraction data is significantly more complex than for small molecules, primarily due to the nature of the phase problem. It generally involves many iterations of calculation, map interpretation and model building. It also relies extensively on input from an expert crystallographer. We have previously proposed that this process could be enhanced through a strategy that integrates research in crystallography and artificial intelligence and rephrases the problem as a hierarchical and iterative scene analysis exercise (Fortier et al., 1993; Glasgow et al., 1993). The goal of such an exercise is to reconstruct and interpret images of a protein at progressively higher resolution. For an initial low-resolution map, where the protein appears as a simple object outlined by its molecular envelope, the goal is to locate and identify protein and solvent regions. At medium-resolution, the goal involves locating amino acid residues along the main chain and identifying secondary structure motifs. At higher resolution, the analysis would attend to the identification of residues and, possibly, the location of individual atoms.

A primary step in any scene analysis (whether vision, medical or crystallographic data are used) is to automatically partition an image representation into disjoint regions. Ideally, each segmented region corresponds to a semantically meaningful component or object in the scene. These parts can then be used as input to a high-level classification task. The processes of segmentation and classification may be interwoven; domain knowledge, in the form of a partial interpretation, is often useful for assessing and guiding further segmentation.

Several approaches to image segmentation and classification have been considered in the vision literature. Of particular interest for the molecular domain is an approach described by Besl and Jain (1986) , where the surface curvature and sign of a Gaussian is derived for each point on the surface of a range image. Image segmentation is then achieved through the identification of primitive critical points (peaks, pits, ridges, etc.). Haralick et al. (1983) defined a similar set of topographic features for use in 2D image analysis, Wang and Pavlidis (1993), and later Lee and Kim (1995), extended this work to extract features for character recognition. Gauch and Pizer (1993) also identify ridge and valley bottoms in 2D images, where a ridge is defined as a point where the intensity falls off sharply in two directions and a valley bottom is a point where the intensity increases sharply in two directions. They further examined the behavior of the ridges and valleys through scale space; the resulting resolution hierarchies could be used to guide segmentation. Maintz et al. (1996) and Guziec and Ayache (1992) use 3D differential operators through scale space to define ridges and troughs. These provide landmarks which can be used to register images. Leonardis, Gupta and Bajcsy (1993, 1995)) use an approach where surface fitting (using iterative regression) and volume fitting (model recovery) are initiated independently; the local-to-global surface along the third axis may be large, and, in such a case, it is necessary to align and interpolate over multiple 2D slices.

fitting is used to guide multiple global-to-local volume fittings, and is used in the evaluation of possible models.

As will be discussed in the next section, a topological approach is being used for the segmentation and classification of molecular scenes. Similar to the method of Gauch and Pizer, critical points are used to delineate a skeletal image of a protein and segment it into meaningful parts. These critical points are considered in the analysis of the segmented parts. This approach can also be compared to a skeletonization method, which has been described by Hilditch (1969) and applied in protein crystallography by Greer (1974) . Unlike Greer's algorithm, which "thins" an electron density map to form a skeleton that traces the main and secondary chains of the molecule, our proposed representation preserves the original volumetric shape information by retaining the curvatures of electron density at the critical points. Furthermore, rather than thinning electron density to form a skeleton, our approach reconstructs the backbone of a protein by connecting critical points into a 3D graph structure.

Critical points in an image have also been considered in the medical domain. Higgins et al. (1996) analyze coronary angiograms from CT data by thresholding to define "bright" regions that correspond to the area around peak critical points. These seed regions are then grown along ridges until there is a steep dropoff. Post-processing removes cavities and spurs. The resulting volume is a tree-like structure, which is then skeletonized and further pruned to provide axes. The axes are converted to sets of line segments with some minimum length. This is similar to BONES (Jones, Zou, Cowan, & Kjeldgaard, 1991), a graphical method which has been developed and applied to the interpretation of medium- to high-resolution protein maps. This method incorporates a thinning algorithm and postprocessing analysis for electron density maps. Also worth mentioning is a previously described methodology for outlining the envelope of a protein molecule in its crystallographic environment (Wang, 1985).

A distinct advantage of molecular scene analysis, over many applications in machine vision, is the regularity of chemical structures and the availability of previously determined molecules in the Protein Data Bank (PDB) (Bernstein, Koetzle, Williams, & Jr., 1977). This database of 3D structures forms the basis of a comprehensive knowledge base for template building and pattern recognition in molecular scene analysis (Conklin, Fortier, & Glasgow, 1993b; Hunter & States, 1991; Unger, Harel, Wherland, & Sussman, 1989); although the scenes we wish to analyze are novel, their substructures most likely have appeared in previously determined structures. Another significant difference between molecular and visual scene analysis is that diffraction data resulting from protein experiments yield 3D images, simplifying or eliminating many of the problems faced in low-level vision (e.g., occlusion, shading). The complexity that does exist in the crystallographic domain relates to the incompleteness of data due to the phase problem.

## 4. Segmentation and Interpretation of Protein Images

The use of artificial intelligence techniques to assist in crystal structure determination, particularly for the interpretation of electron density maps, was first envisioned by Feigenbaum, Engelmore and Johnson (1977) and pursued in the *Crysalis* project (Terry, 1983). In conjunction with this earlier project, a topological approach to the representation and analysis of protein electron density maps was implemented in a program called ORCRIT (Johnson, 1977). Recent studies suggest that this approach can also be applied to the segmentation of medium-resolution protein images (Leherte, Baxter, Glasgow, & Fortier, 1994a; Leherte, Fortier, Glasgow, & Allen, 1994b). In this section we describe and further support the feasibility of a topological approach for the analysis of low and medium-resolution electron density maps of proteins.

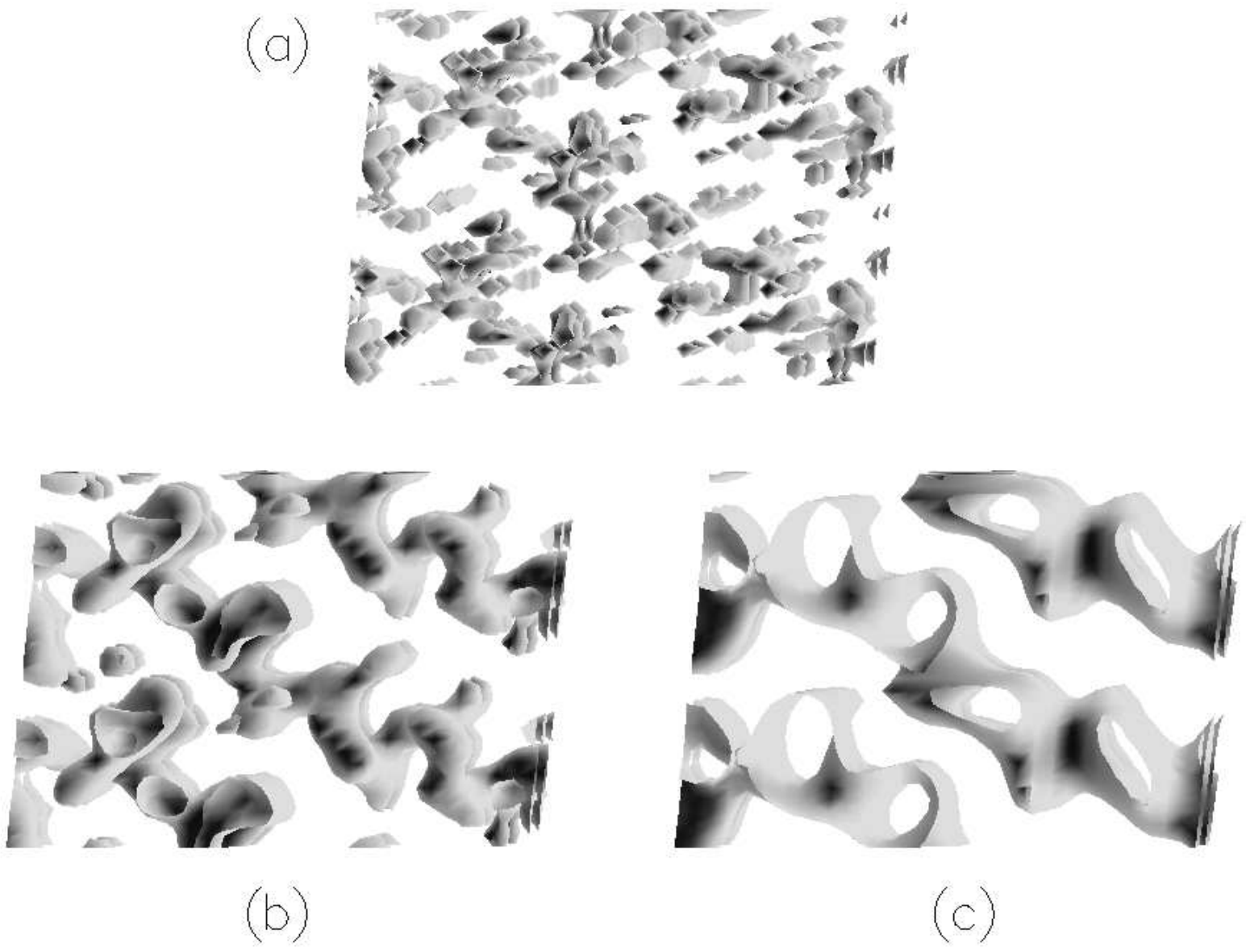

Figure 4: Depictions of electron density maps viewed at (a) 1 Å, (b) 3 Å, and (c) 5 Å resolution

The information stored in an electron density map for a protein may be represented and analyzed at various levels of detail (see Figure 4)[4]. At high-resolution (Figure 4(a)) individual atoms are observable; at medium-resolution (Figure 4(b)) atoms are not observable, but it is possible to differentiate the backbone of the protein from its side chains and secondary structure motifs may be discernerable. A clear segmentation between the protein molecular envelope (region in which the atoms of the protein reside) and the surrounding solvent region can still be seen at low-resolution (Figure 4(c)).

4. Resolution in the electron density images of proteins is often measured in terms of angstrom (Å) units, where 1 Å=$10^{-10}$ meters.

Methods from both machine vision and crystallography were considered in the development of our computational approach to the analysis of protein structures. Among those studied, a topological analysis provided the most natural way to catch the fluctuations of the density function in the molecular image. In this section we overview such a method for mapping a 3D electron density map onto a graph that traces the backbone of the protein structure. We present results of applying the method for the segmentation of low and medium-resolution maps of protein structures reconstructed using the Protein Databank of Brookhaven. As well, we show how critical point graphs constructed for medium resolution maps can be further analyzed in order to identify $\alpha$-helix and $\beta$-sheet substructures.

### 4.1 Representation of Protein Structure

Crucial to a molecular scene analysis is a representation that will capture molecular shape and structure information at varying levels of resolution; an important step in a molecular scene analysis is the parsing of a protein, or protein fragments, into shape primitives so as to allow for their rapid and accurate identification. Shape information can be extracted from several depictive representations – for example, van der Waals surfaces, electrostatic potentials or electron density distributions. Since molecular scene analysis is primarily concerned with images reconstructed from crystallographic experiments, electron density maps provide a natural and convenient choice for the input representation.

As mentioned earlier, an electron density map is depicted as a 3D array of real, non-negative values corresponding to approximations of the electron density distribution at points in the unit cell of a crystal. For the task of segmenting this map into meaningful parts, we also consider the array in terms of a smooth and continuous function $\rho$, which maps an integer vector $r = (x, y, z)$ to a value corresponding to the electron density at location $r$ in the electron density map. Similar to related formalisms in vision[5], the information in an electron density map is uninterpreted and at too detailed a level to allow for rapid computational analysis. Thus, it is essential to transform this array representation into a simpler form that captures the relevant shape information and discards unnecessary and distracting details. The desired representation should satisfy the three criteria put forward by Marr and Nishihara concerning: 1) *accessibility* – the representation should be derivable from the initial electron density map at reasonable computing costs; 2) *scope and uniqueness* – the representation should provide a unique description of all possible molecular shapes at varying levels of resolution; and 3) *stability and sensitivity* – the representation should capture the more general (less variant) properties of molecular shapes, along with their finer distinctions.

Several models for the representation and segmentation of protein structures were considered. These included a generalized cylinder model (Binford, 1971), and a skeletonization method (Greer, 1974; Hilditch, 1969). We choose a topological approach, which has been previously applied in both chemistry and machine vision. In chemistry, the approach has proven useful for characterizing the shape properties of the electron density distribution through the location and attributes of its *critical points* (points where the gradient of the electron density is equal to zero) (Johnson, 1977).

5. The level of representation of the electron density map is comparable to a 3D version of the primal sketch in Marr and Nishihara's formalism (1978) .

In the following section, we describe the representation of a protein structure in terms of a set of critical points obtained through a topological analysis.

### 4.2 Deriving Critical Point Graphs

In the proposed topological approach to protein image interpretation, a protein is segmented into its meaningful parts through the location and identification of the points in the electron density map where the gradients vanish (zero-crossings). At such points, local maxima and minima are defined by computing second derivatives which adopt negative or positive values respectively. The first derivatives of the electron density function $\rho$ characterize the zero-crossings, and the second derivatives provide information on the zero-crossings; in particular, they identify the type of critical points for the map. For each index value $r = (x, y, z)$ in the electron density map, we define a function $\rho(r)$.

Candidate grid points (those that are a maximum or a minimum along three mutually orthogonal axes) are chosen and the function $\rho(r)$ is evaluated in their vicinity by determining three polynomials (one along each of the axes) using a least square fitting. $\rho(r)$ is the tensor product of these three polynomials. The location of a critical point is derived using the first derivative of $\rho(r)$. The second derivatives are used to determine the nature of the critical point. For each critical point, we construct a Hessian matrix:

$$\mathbf{H}(\mathrm{r}) = \begin{vmatrix} \partial^2\rho/\partial x^2 & \partial^2\rho/\partial x\partial y & \partial^2\rho/\partial x\partial z \\ \partial^2\rho/\partial y\partial x & \partial^2\rho/\partial y^2 & \partial^2\rho/\partial y\partial z \\ \partial^2\rho/\partial z\partial x & \partial^2\rho/\partial z\partial y & \partial^2\rho/\partial z^2 \end{vmatrix}$$

This matrix is then put in a diagonal form in which three principal second derivatives are computed for the index value $r$:

$$\mathbf{H'}(\mathrm{r}) = \begin{vmatrix} \partial^2\rho/\partial(x')^2 & 0 & 0 \\ 0 & \partial^2\rho/\partial(y')^2 & 0 \\ 0 & 0 & \partial^2\rho/\partial(z')^2 \end{vmatrix}$$

The three non-zero diagonal elements of array $\mathbf{H'}(\mathrm{r})$ – the eigenvalues – are used to determine the type of critical points of the electron density map. Four possible cases are considered depending upon the number of negative eigenvalues, $n_E$. When $n_E = 3$, the critical point $r$ corresponds to a local maximum or *peak*; a point where $n_E = 2$ is a saddle point or *pass*. $n_E = 1$ corresponds to a saddle point or *pale*, while $n_E$=0 characterizes $r$ as a *pit*. Figure 5 depicts a 2D graphical projection of potential critical points.

High density peaks and passes are the only critical points currently being considered in our study. Low density critical points are less significant since they are often indistinguishable from noise in experimental data.

The use of the critical point mapping as a method for analyzing protein electron density maps was first proposed by Johnson (1997) for the analysis of medium to high-resolution protein electron density maps. Within the framework of the molecular scene analysis project, the use of critical points is being extended for the analysis of medium and low-resolution maps of proteins. The topological approach to the segmentation of proteins was initially implemented by Johnson in the computer program ORCRIT (Johnson, 1977). By first locating and then connecting the critical points, this program generates a representation that captures the skeleton and the volumetric features of a protein image. The occurrence probability of a connection between two critical points $i$ and $j$ is determined by following the density gradient vector $\nabla\rho(\mathbf{r})$. For each pair of critical points, the program calculates a weight $w_{ij}$, which is inversely proportional to the occurrence probability of the connection. The collection of critical points and their linkage is represented as a set of minimal spanning trees (connected acyclic graphs of minimal weight)[6].

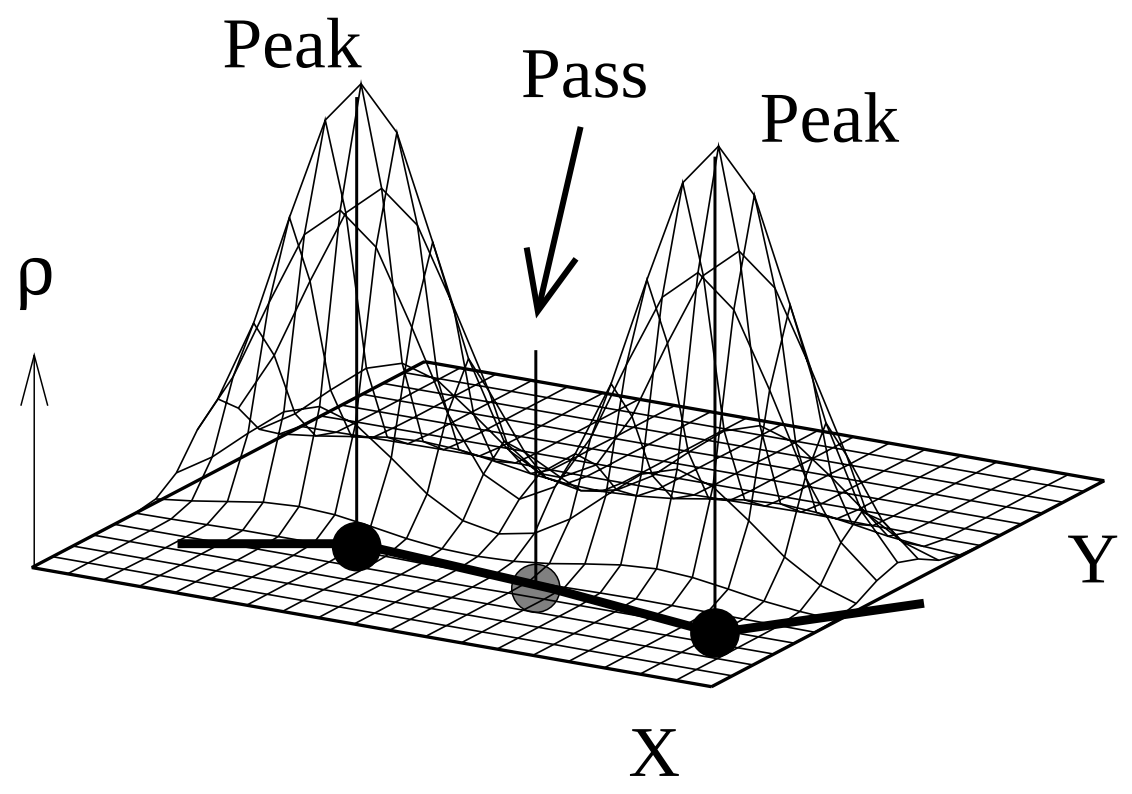

Figure 5: Graphical illustration of critical points in 2D (X and Y) plotted against density function $\rho$.

### 4.3 Results of Segmentation at Medium and Low Resolution

This section presents experimental studies that have been carried out on electron density maps at 3 Å resolution. Computations were first performed on calculated maps reconstructed from available structural data in order to generate a procedure for the further analysis of experimental maps. Three protein structures: *Phospholipase A2* (1BP2), *Ribonuclease T1 complex* (1RNT) and *Trypsin inhibitor* (4PTI), retrieved from the PDB (Bernstein et al., 1977), were considered. These structures are composed of 123, 104 and 53 amino acid residues, respectively, and were chosen because of the quality of the data sets. The electron density images for the proteins were constructed using the XTAL program package (Hall & Stewart, 1990), and were then analyzed using a version of ORCRIT which was extended and enhanced to construct and interpret critical point graphs for low- and medium-resolution electron density maps.

Applying ORCRIT to electron density maps generated at medium resolution provides for a detailed analysis of the protein structures (Leherte et al., 1994b). As illustrated in Figure 6, the topological approach produces a skeleton of a protein backbone as a sequence

6. As will be discussed in Section 4, this part of the program is currently being modified to allow for cycles in the graph.

of alternating peaks (dark circles) and passes (light circles). The results obtained from the analysis of calculated electron density maps at 3 Å resolution led to the following observations:

- A peak in the linear sequence is generally associated with a single residue of the primary sequence for the protein.

- A pass in the sequence generally corresponds to a bond or chemical interaction that links two amino acid residues (peaks).

Thus, the critical point graph can be decomposed into linear sequences of alternating peaks and passes corresponding to the main chain or backbone of the protein. For larger residues, side chains may also be observed in the graph as side branches consisting of a peak/pass motif. These observations are featured in Figure 7, which illustrates a critical point graph and an electron density contour for the unit cell of a protein crystal.

In practice, we found that the critical point graph included some arcs originating from the presence of connections between critical points associated with non-adjacent residues. This is illustrated in Figure 6 and in the bottom left corner of Figure 7; in the main chain of both graphs there is a jump that occurred as a result of a disulfide bridge between nearby residues. These bonds can often be identified, however, through further analysis of the critical point graphs. For example, disulfide bridges are discerned from the representation through the higher density values of their associated cysteine residues. To overcome the problem of errors in the critical point graph due to ambiguous data we plan to generate multiple possible models for a protein, corresponding to different hypothesized backbones resulting from the critical point graph. Thus, several alternative hypotheses will be considered and used to refine the image in an iterative approach to scene analysis.

Experiments were also carried out at low (5 Å) resolution, where the following was observed:

- Secondary structure motifs, such as $\alpha$-helices and $\beta$-strands, correspond to linear (or quasi-linear) sequences of critical points. In the case of helices, these sequences trace the central axis of the structure (see Figure 8), whereas they tend to catch the backbone itself for $\beta$-strands. The average distance between observed critical points and a model of the protein backbone was $1.68 \pm 0.59$ Å.

- Other non-linear sequences of critical points were sometimes found to be representative of irregular protein motifs such as loops.

- Highly connected, small graphs of critical points appear in other regions of the maps: in the solvent region, at the disulfide bridges and between close protein segments.

Although the results of segmenting protein images using the topological approach has proved promising, there is ongoing research being carried out to improve and enhance ORCRIT. In particular, we are redeveloping the code for building the graph of critical points. The new version of the code will incorporate more domain knowledge in order to find multiple possible backbone traces. It will also incorporate a spline (rather than polynomial) fitting function to interpolate the critical points for constructing the function $\rho(r)$.

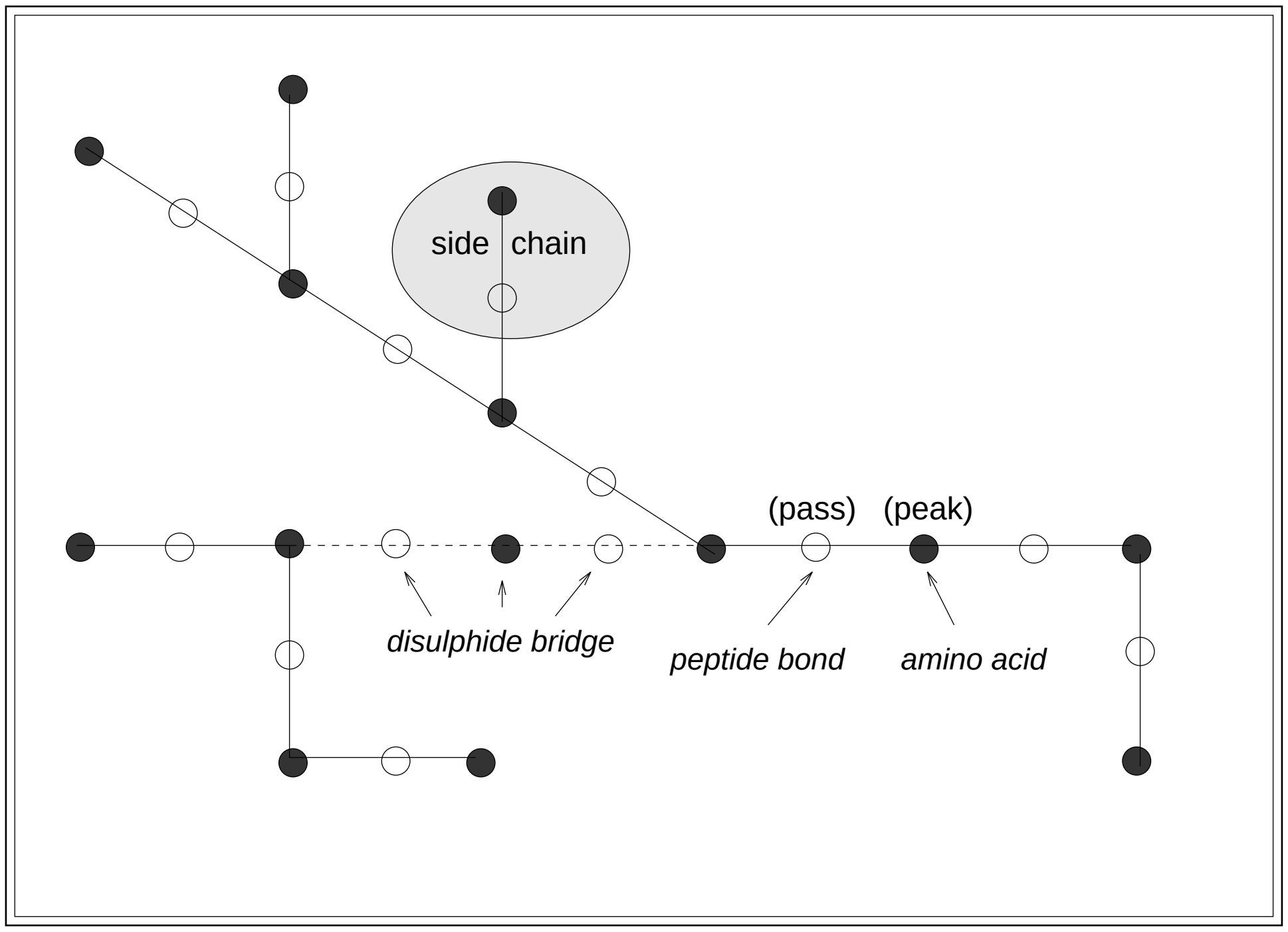

Figure 6: Planar representation of a critical point sequence obtained from a topological analysis of an electron density map at 3 Å resolution.

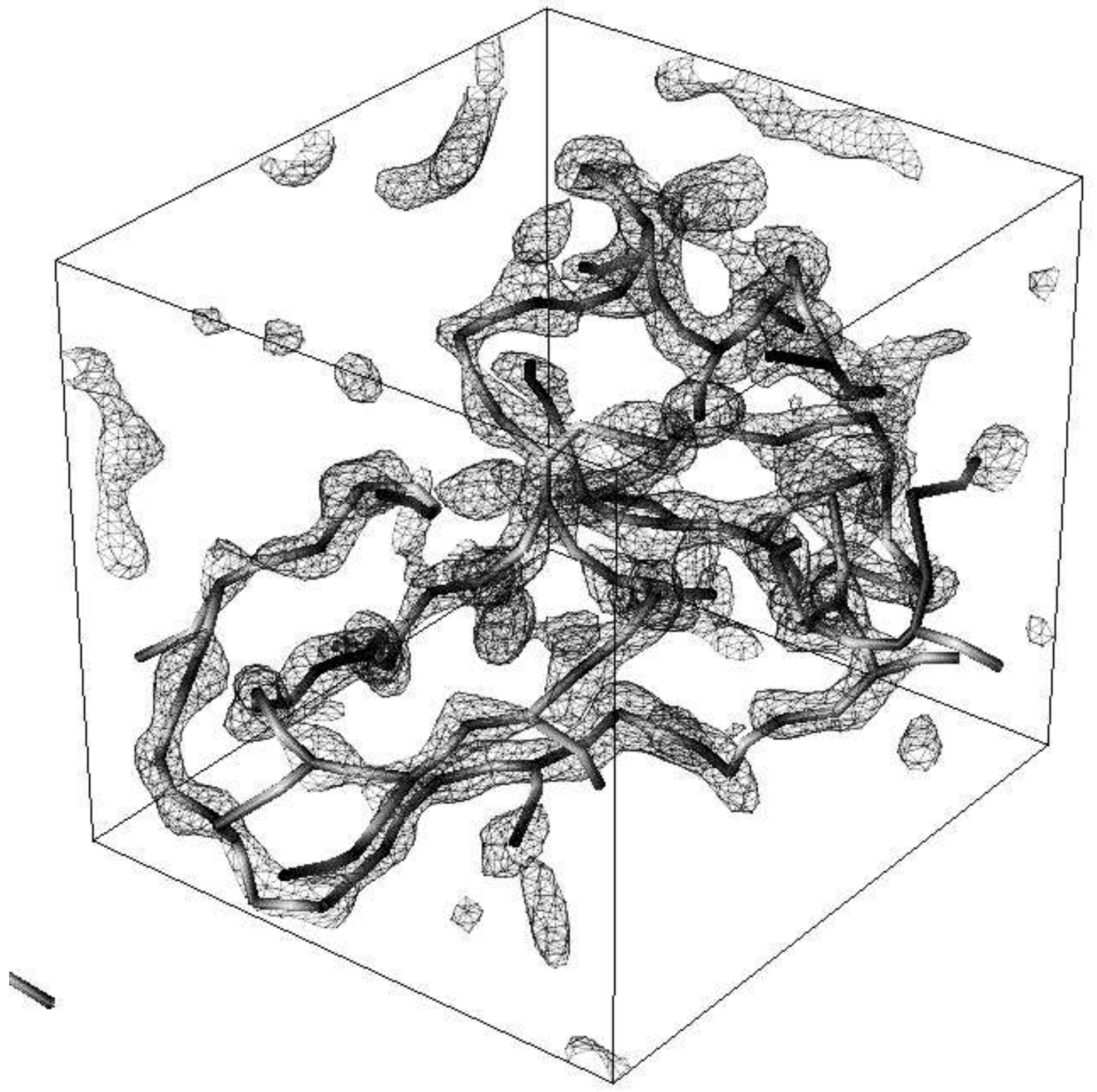

Figure 7: 3D contour and critical point graph for a unit cell of protein 4PT1 (58 residues) constructed at 3 Å resolution. The critical point graph in this figure was generated using the output of the ORCRIT program.

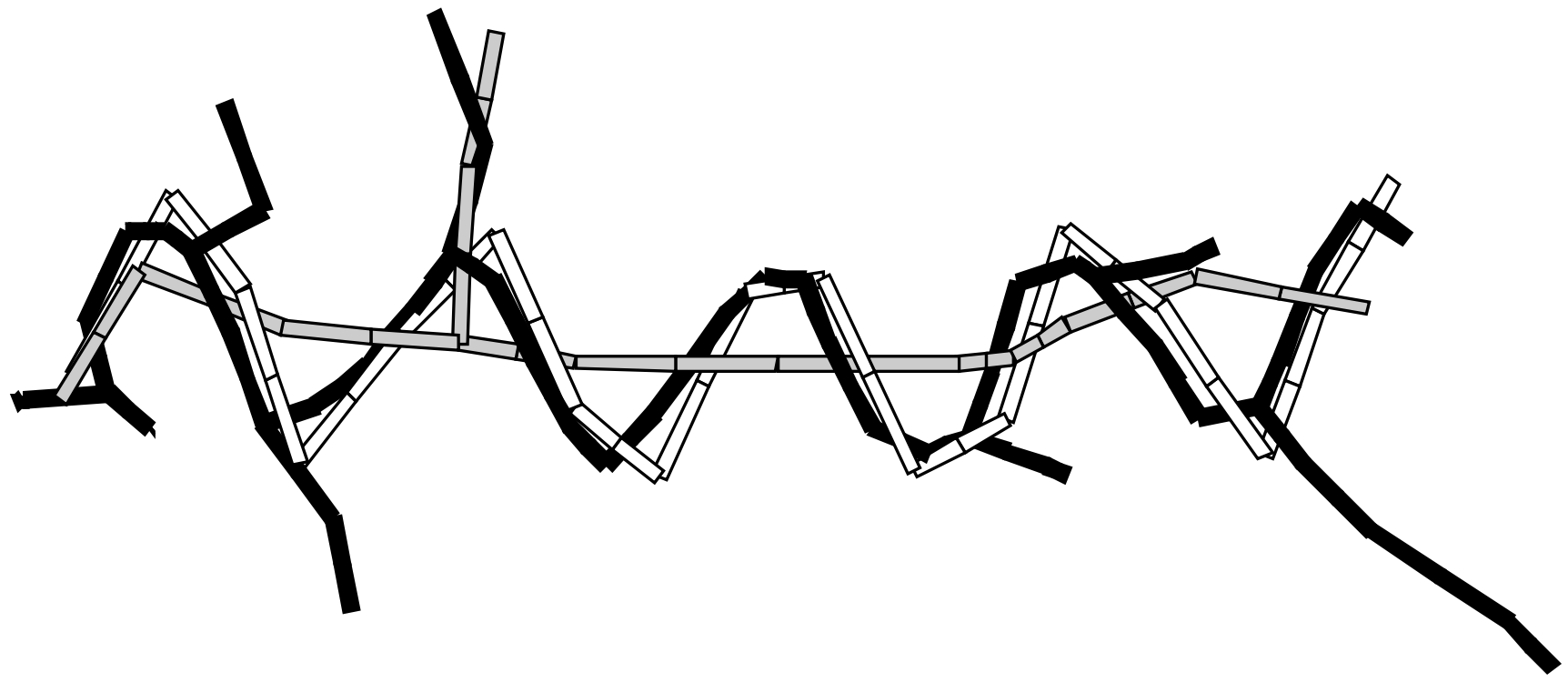

Figure 8: Critical point graph of $C_\alpha$ chain (black) for an helix motif at 3 Å (white) and 5 Å (gray) resolution.

Currently, we are investigating the relationships between critical points at varying resolutions. Figure 8 illustrates the superimposition of critical point representations of an helix motif at low and medium resolution obtained using ORCRIT. As this figure shows, the linear segment of critical points derived at 5 Å resolution approximates the more detailed critical point graph of a helix derived at 3 Å. More careful analysis suggests that there exists a hierarchy between the peaks and passes at 5 Å and those at 3 Å. This relationship between critical points at medium and low-resolution is illustrated in Figure 9, where individual critical points at 5 Å resolution are associated with individual or multiple points at medium-resolution.

## 5. Methods for Secondary Structure Identification

Once a critical point graph is constructed, it can be analyzed to classify substructures in the protein. A statistical analysis of the conformation of critical point sequences in terms of geometrical parameters of motifs consisting of four sequential peak critical points ($p_i, p_{i+1}, p_{i+2}, p_{i+3}$) suggests that the most useful parameters for the identification of $\alpha$-helices and $\beta$-strands are the distance between peaks $p_i$ and $p_{i+3}$, and the individual and planar angles among critical points. We describe how these criteria were used to formulate rules for the identification of secondary structure motifs in a medium-resolution electron density map of a protein.

In a previous paper (Leherte et al., 1994a), we described an approach to secondary structure identification in which the geometrical constraints in a critical point subgraph were treated in a boolean fashion, i.e., they were either satisfied or not depending upon whether or not their values fell within predetermined ranges. The procedure described in the current paper relies on probabilistic belief measures based on statistics derived from the PDB. Following, we describe alternative approaches based on the comparison of critical point graphs with idealized secondary structure motifs. The templates we used consisted of local subsequences of critical points, each point denoting a residue in an idealized model of a protein.

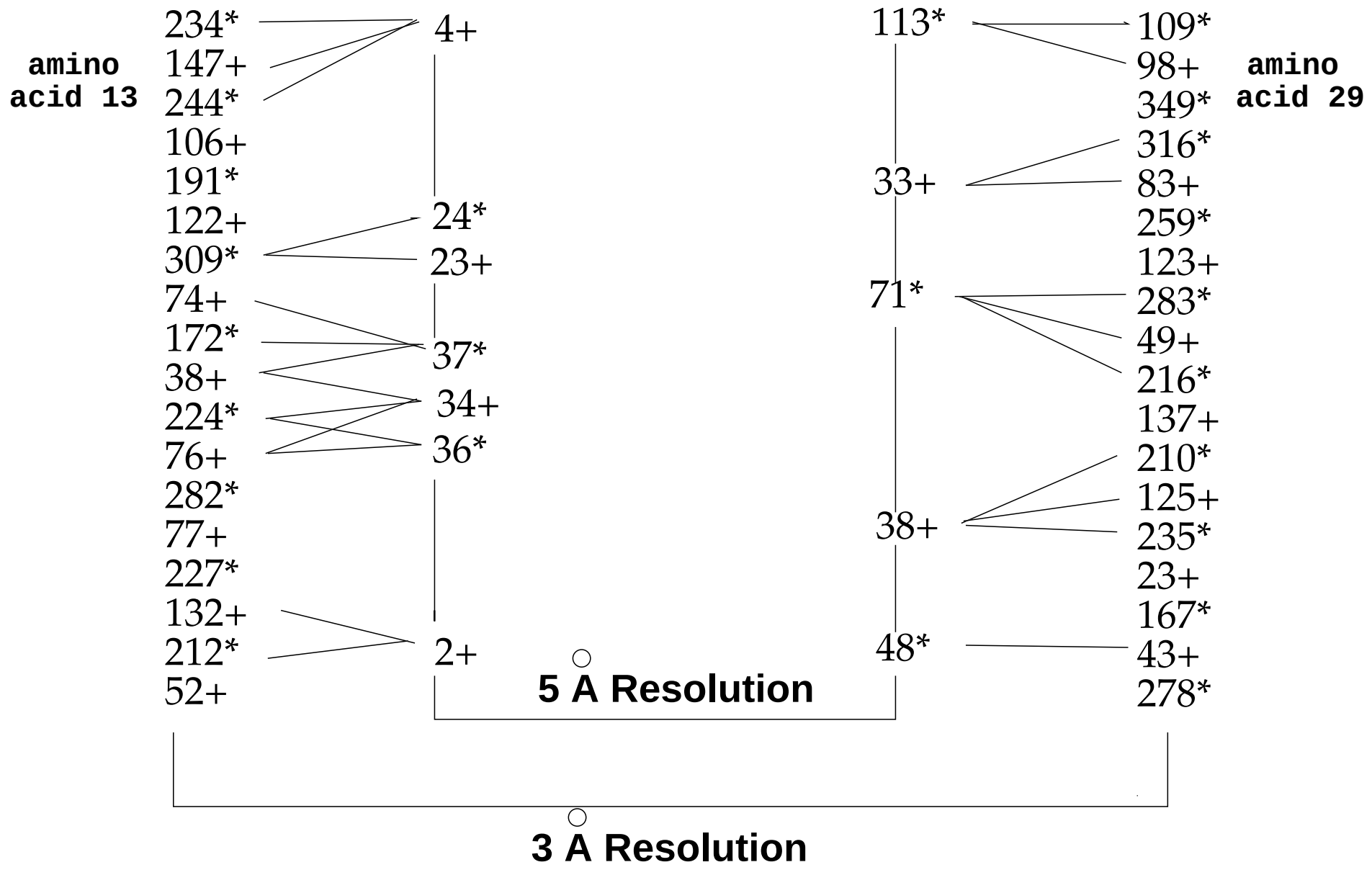

Figure 9: Relationship between critical points at 3 and 5 Å resolution for amino acids 13 to 29 in protein structure 1RNT. '+' and '*' denote peaks and passes in the main chain. The numeric values correspond to the critical points location in an ordered list (based on electron density) of all points. The correspondence between points at different levels in the hierarchy is based on their interdistance (must be less than 2 Å).

### 5.1 Estimating Probabilities and Combining Evidence

To construct our model templates, we first performed a statistical analysis of 63 non-homologous protein structures retrieved from the PDB. A set of occurrence probability distributions $f(ssm|g)$ were estimated for given secondary structure motifs ($ssm$) and geometric constraints ($g$). We derived these values for the $\alpha$-helix, and $\beta$-sheet and turn motifs and for geometric constraints based on torsion and bond angles and on relative distance between residues.

In building procedures for structure recognition and discrimination, two important issues must be faced: first, how does one compute the primitive marginal and conditional probability estimates such as $f(ssm|g)$; and, second, how does one *combine* the information from the several different pieces of geometric evidence for or against each class, as in $f(ssm|g_1, g_2)$? There are many tradeoffs to consider.

For the single-attribute probability estimations, $f(ssm|g)$, one can use discrete categories and estimate probabilities from frequency counts. If only a few "bins" are employed for, e.g., ranges of critical point inter-distance values, then achieving sufficient sample sizes for bin counts presents little difficulty. As the number of bins grows and the "width" of each bin correspondingly shrinks, the resulting histogram becomes a better and better approximator of an underlying continuous distribution, but here problems with small counts lead to larger sampling error (variance). If one chooses to fit continuous distributions to the data, the bias/variance dilemma manifests itself in the choice of distribution types and the number of parameters for parameterized models.

In combining the individual terms representing secondary structure evidence, one must address the issue of inter-attribute dependencies and the accuracy and efficiency tradeoffs it poses. Put most simply, how do we compute $f(ssm|g1, g2)$ from $f(ssm|g1)$ and $f(ssm|g2)$? A pure Bayesian approach without underlying independence assumptions requires exponentially many terms, and we therefore seek heuristic shortcuts.

Two methods for determining confidence values for secondary structure assessment were studied and applied to the problem of secondary structure identification: 1) a Bayesian, or Minimum Message Length (MML) approach, similar to that previously used in protein substructure classification in (Dowe, Allison, Dix, Hunter, Wallace, & Edgoose, 1996), and 2) an approach similar to that used in expert systems such as MYCIN (Shortliffe, 1976). We should emphasize here that the primary goal of our described research is the construction of effective structure recognition systems, rather than the comparison of alternative methods of machine learning and classification *per se*.

### 5.2 A Bayesian/MML Approach

In adopting a Bayesian latent class analysis approach to the problem, we decided to treat the estimation and combination issues together by fitting mixtures of continuous distributions to the data for each class, under the conditional independence assumption commonly used in mixture model approaches to classification and clustering (McLachlan & Basford, 1988).

In a latent class analysis approach to finding structure in a set of datapoints, one begins with an underlying parameterized model. For example, one might posit that a set of points represented by a 2D scatterplot was generated by a 2D Gaussian (normal) distribution, with means $\mu_1, \mu_2$ and covariance matrix $V_{12}$. Or the data might be better explained by a mixture, or weighted sum, of several Gaussian distributions, each with its own 2D mean vector and covariance matrix. In this approach, one tries to find an optimal set of parameter values for the representation of each datapoint $\bar{x} = (x_1, x_2)$. Optimality may be defined in terms of maximum likelihood, Bayes optimality, or, as in our case, minimum message length (MML)[7].

In the case at hand, we have 11 dimensions instead of 2, and not all of the dimensions are best modeled by simple Gaussians. It is generally accepted that angular data should be modeled by one of the circular distributions such as the von Mises distribution (Fisher, 1993).

Two independence assumptions, crucial to computational efficiency and data efficiency, underlie this approach:

1. Within a given class, the attributes characterizing a segment are mutually independent.

2. The separate datapoints, each corresponding to a segment, are mutually independent.

Although neither of these assumptions is strictly true in this application, these assumptions are commonly made in such circumstances, and the methods based on them work well in practice. Where dependence on these assumptions proves untenable, one can employ more elaborate models that incorporate explicit dependencies, such as Bayes Nets and graphical models (Buntine, 1994).

### 5.3 A MYCIN-Like Approach

We determined that a method similar to the one used for the diagnosis system MYCIN (Rich & Knight, 1991) was also effective for our application. Frequency distribution values provide measures of belief and disbelief for secondary structure assignments based on individual geometric constraints:

$$MB(ssm, g) = f(ssm, g) \tag{1}$$

$$MD(ssm, g) = f(not_ssm, g) \tag{2}$$

where $MB(ssm, g)$ is a measure of belief in the hypothesis that a given peak is associated with a secondary structure $ssm$ given the evidence $g$, whereas $MD(ssm, g)$ measures the

7. However, we use the general term "Bayesian" informally for Bayesian, Minimum Description Length and MML approaches to distinguish them jointly from other, especially "frequentist" approaches.

the extent to which evidence $g$ supports the negation of hypothesis $ssm$ for the peak. Figure 10 illustrates the computed probability distributions for each geometric constraint (bond angle, distance and torsion angle) and each secondary structure motif (helix, strand).

Like the modified Bayesian mixture modeling approach described in the previous section, the MYCIN methodology represents another heuristic approximation to a pure "naive" Bayes approach. In this case, the initial primitive probability terms are simple frequency counts and the rules for combining probabilities assume implicitly that different evidence sources are independent. This can be shown to lead to nonsensical classifications in extreme cases, though in practice the approach often works quite well.

When two or more pieces of evidence are considered, the confidence measures are computed using the following formulae:

$$MB(ssm, g_1 \wedge g_2) = MB(ssm, g_1) + MB(ssm, g_2)(1 - MB(ssm, g_1)) \quad (3)$$

$$MD(ssm, g_1 \wedge g_2) = MD(ssm, g_1) + MD(ssm, g_2)(1 - MD(ssm, g_1)) \quad (4)$$

Given these measures, an overall certainty factor, $CF$, can be determined for each peak $p$ in the critical point graph as the difference between the belief and disbelief measures:

$$CF(ssm, g) = MB(ssm, g) - MD(ssm, g) \quad (5)$$

where $g$ corresponds to the geometric constraints associated with peak $p$.

A result of the application of this method to the interpretation of an ideal critical point graph is illustrated in Figure 11. This graph shows that broad bands of positive $CF$s are indeed representative of regular secondary structure motifs such as $\alpha$-helices and $\beta$-strands.

In practice, for each critical point, our approach constructs a CF value for each secondary structure hypothesis. At the end of the procedure, the hypothesis with the highest CF value is selected. The final results are thus a set of sequences of CF values characterized by subsequences of various lengths with identical secondary structure selection.

When two or more pieces of evidence are considered, the confidence measures are computed using the following formulae:

$$MB(ssm, g_1 \wedge g_2) = MB(ssm, g_1) + MB(ssm, g_2)(1 - MB(ssm, g_1)) \quad (6)$$

$$MD(ssm, g_1 \wedge g_2) = MD(ssm, g_1) + MD(ssm, g_2)(1 - MD(ssm, g_1)) \quad (7)$$

Given these measures, an overall certainty factor, $CF$, can be determined for each peak $p$ in the critical point graph as the difference between the belief and disbelief measures:

$$CF(ssm, g) = MB(ssm, g) - MD(ssm, g) \quad (8)$$

where $g$ corresponds to the geometric constraints associated with peak $p$. A result of the application of this method to the interpretation of an ideal critical point graph is illustrated in Figure 11. This graph shows that broad bands of positive $CF$s are indeed representative of regular secondary structure motifs such as $\alpha$-helices and $\beta$-strands.

### 5.4 Results

Following we demonstrate how the two methods of analysis described in the previous section can be applied to the identification of secondary structure in critical point graphs. We consider graphs constructed from both ideal and experimental electron density maps.

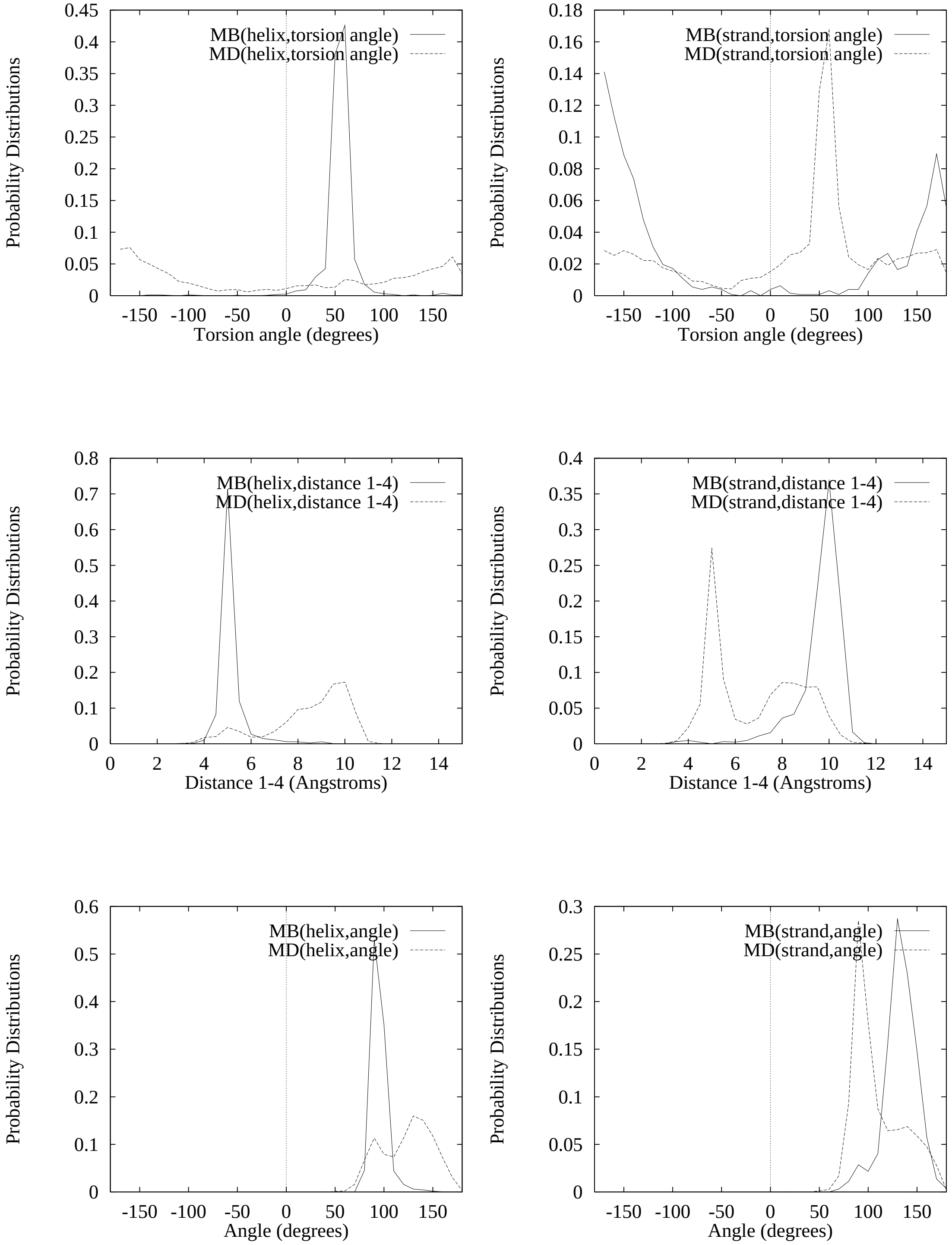

Figure 10: Probability distributions computed for measures of belief (MB) and disbelief (MD) for a given secondary structure motif and geometric constraint.

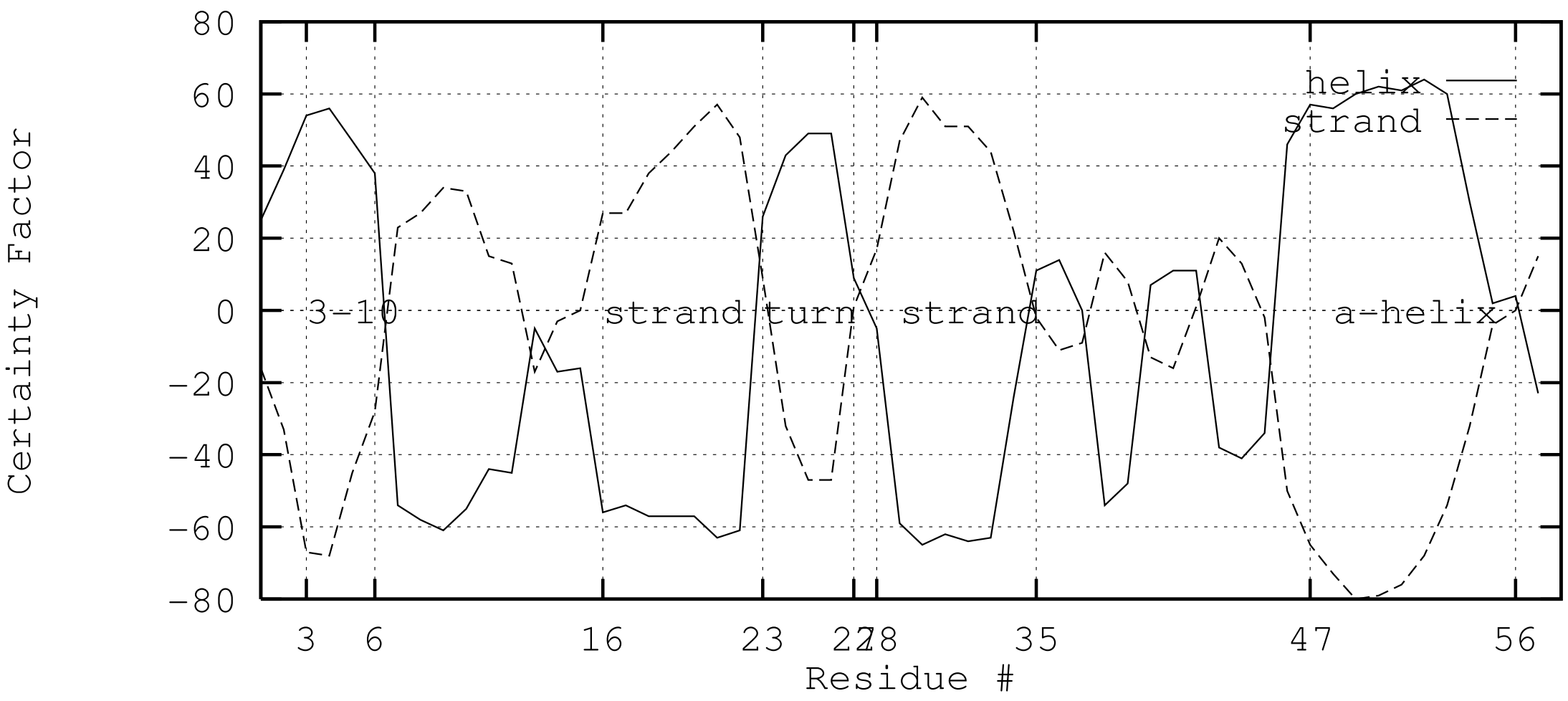

Figure 11: Certainty factors obtained for $\alpha$-helix and $\beta$-strand hypotheses for an ideal critical point representation of protein 4PTI (58 residues) at medium-resolution. The figure also denotes the correct interpretation for residues 16-23 (strand), 23-27 (turn), 28-35 (strand) and 47-56 (helix).

### 5.5 Application to Ideal Data

Two Bayesian/MML and one MYCIN-based analyses were applied to the secondary structure identification of ideal critical point trees. The first Bayesian module ($\text{Bayes}_1$) was trained using 60 out of the 63 ideal protein structure representations that were previously used to generate occurrence frequency distribution functions for torsion angles, distances and planar angles (Figure 10). The second Bayesian module ($\text{Bayes}_2$) was trained using 46 ideal critical point trees and 18 critical point representations obtained from the ORCRIT analysis of 3 Å resolution reconstructed electron density maps. Three ideal critical point representations were kept for testing: *Cytochrome C2* (2C2C – 112 residues) is characterized by helices and turns only; *Penicillopepsin* (3APP – 323 residues) contains short helices (8 residues or less), turns, and $\beta$-strands up to 14 residues long; and the photosynthetic reaction centre of *Rhodobacter Sphaeroides* (4RCR – 266 residues) is a rich $\alpha$-helix structure with regular segments of up to 24 residues.

The statistical Bayesian modules allow the classification of critical points for which 11 geometrical attributes can be calculated: except for the first three and the last three points of a critical point sequence, all points participate in four torsion angles, four distances ($p_i$ to $p_{i+3}$), and three planar angles. The Bayesian module can thus be applied to the secondary structure recognition of segments which contain 7 peak critical points or more, while the MYCIN-based module is applicable to 4-point (or longer) sequences. However, for comparison purposes, only points for which all 11 geometrical attributes could be calculated were considered for recognition. Results are presented in Table 1. This table reports the number of segments[8] which have been correctly or incorrectly identified by either the Bayesian approaches or the MYCIN-based module. All modules were designed to classify recognized secondary structure motifs among four different classes: 'helix', 'strand', 'turn', and 'other'. Regarding the class 'helix', a distinction between $\alpha$-helices and helices $3_{10}$ was made *a posteriori* to help in the interpretation of the results.

Two types of percentage values are given in Table 1. The first type, i.e., the percentage of actual secondary structure information that was identified, was computed over the total number of actual secondary structure motifs present in the three test protein structures: 25 $\alpha$-helices, 25 $\beta$-strands, 10 $3_{10}$ helices, and 42 turns. Higher percentage values observed for the MYCIN-based results versus the Bayesian results come from the fact that longer segments are recognized as potential helices or strands. A better overlap with the actual secondary structure motifs is thus more likely to occur using the MYCIN approach. This is illustrated by the first two examples described in Figure 12. Selected secondary structure motifs of proteins 2C2C and 3APP are compared with the hypotheses generated by the MYCIN-based and Bayes modules. It is observed that, in these two cases, the longer identifications provided using MYCIN are closer to the actual secondary structure of the amino acid sequences.

The percentage of correctly identified points within the ideal critical point segments was computed over the total number of assigned critical point segments reported in Table 1. Regarding the class 'helix', the longer segments discovered by the MYCIN-based module, as well as the larger number of incorrectly recognized segments, lead to lower percentage values for this particular method. This is shown in the third example displayed in Figure 12. The MYCIN-based module associates a long $\alpha$-helix with this particular amino acid sequence of protein 4RCR which deviates from ideality by five residues, while the Bayes modules provides reasonable solutions.

It is worthwhile to not that even if all of the segments are correctly assigned, the percentage of correctly identified peaks is not 100%. This is due to the fact that most of the recognized segments (sequences of peaks) are shifted by one residue with respect to the definition given in the PDB file.

From the results reported in Table 1, it is clear that the first Bayesian module allows a finer differentiation between helices and turns (turns are four or five residue long segments whose geometry may be similar to the helix geometry) than the MYCIN-based approach. The MYCIN-based approach tends to assign a label 'helix' to actual turns as shown in Example (4) in Figure 12. On the other hand, $3_{10}$ helices are correctly identified by the MYCIN-derived rules, but less often discerned using the first Bayesian approach (See Example (5) in Figure 12). The MYCIN-based module actually has a strong tendency to exaggerate the helical character of a segment that is distorted with respect to the ideal case. This raises identification ambiguities for 27 (51-24) short segments. No wrong identification is made using the first Bayesian approach, except for one $\beta$-strand. This segment was also identified as a possible strand using the MYCIN-based module, but the hypothesis was later rejected by a post-processing stage which checks for parallelism with other discovered $\beta$-strands.

8. In this table a segment denotes a sequence (length $\geq 2$) of adjacent peak critical points which belong to the same secondary structure class (helix, $\beta$-strand, turn). When comparing results, it should be noted that the PDB data set is, itself, not strictly consistent since varying secondary structure definitions and assignment methods are used to assess the structure of proteins.

| | MYCIN | $Bayes_1$ | $Bayes_2$ |
|---|---|---|---|
| $\alpha$-Helices (actual no = 25) | | | |
| No of (correctly) assigned segments | (24) 51 | (24) 24 | (21) 22 |
| % of correctly recognized actual motifs | 98 | 87 | 88 |
| % of correctly identified peaks | 63 | 83 | 82 |
| $\beta$-Strands (actual no = 25) | | | |
| No of (correctly) assigned segments | (24) 24 | (20) 21 | (24) 30 |
| % of correctly recognized actual motifs | 89 | 71 | 84 |
| % of correctly identified peaks | 82 | 82 | 81 |
| $3_{10}$ Helices (actual no = 10) | | | |
| No of (correctly) assigned segments | (10) 10 | (7) 7 | (7) 7 |
| % of correctly recognized actual motifs | 97 | 56 | 47 |
| % of correctly identified peaks | 70 | 70 | 61 |
| Turns (actual no = 42) | | | |
| No of (correctly) assigned segments | (4) 4 | (12) 12 | (21) 28 |
| % of correctly recognized actual motifs | 7 | 34 | 41 |
| % of correctly identified peaks | 46 | 77 | 59 |

Table 1: Results from the application of two Bayesian approaches and a MYCIN-based method to the recognition of secondary structure motifs in ideal protein backbones constructed from $C_\alpha CO$ centres of mass. Note that the numbers in brackets denote the number of correctly assigned, versus total number of assigned, segments (sequences of peaks).

The application of the second Bayesian approach trained with more realistic critical point representations generated a larger number of identified $\beta$-strands and turns. This however went with a number of incorrect identifications which are, in the case of $\beta$-strands, all associated with very short segments (2 or 3 points). In the case of turns, the percentage of correctly identified critical points is lower (59% with respect to 77%) due to one particular motif containing seven points.

The analysis of ideal critical point trees allows to conclude that the second Bayesian module is more accurate in detecting $\beta$-strand and turn structures (there is an increased number of recognized motifs); but the use of noisy data in the training stage leads to a less acute ability of the module to distinguish short helix-like motifs (there is an increased number of incorrectly identified motifs).

### 5.6 Application to Experimental Data

Above we presented results obtained in applying methods for secondary structure identification to critical point graphs constructed from ideal electron density maps. Following, we describe an application of our methods to the recognition of motifs in a critical point representation constructed by applying ORCRIT to an electron density map generated from experimental data. We also show how our structure recognition approaches can be improved through a postprocessing analysis of the representation. The experiment was carried out using a 3 Å resolution experimental map of *Penicillopepsin*, a protein composed of 323 amino acid residues (Hsu, Delbare, James, & Hofmann, 1977; James & Sielecki, 1983).

Neglecting the passes located between the peaks, geometrical parameters were computed for short fragments composed of seven adjacent peaks in the main branch of the graph for the protein. Before achieving this geometrical analysis, some preprocessing work was done in order to fit the critical point graphs to an ideal model as described above. Distances were computed for sets of adjacent peaks, and peaks separated by a distance smaller than 1.95 Å were merged into a single point situated at their center of mass. The critical point linkage was then checked: if two adjacent peaks were separated by a distance $\leq 5$ Å then the peaks were assumed to be connected. Considering connected sequences of three peaks at a time, if the distance between the first and third peak was smaller than 4 Å, then the middle peak was not considered to be part of the backbone of the protein (i.e., the middle peak probably denotes a side chain). Finally, all resulting sequences of peaks (which are thus likely to be representative of the protein backbone) were submitted to our three secondary structure analysis methods.

Initially poor results observed from the MYCIN-like method motivated the development of a post-processing procedure which was imposed to eliminate all helical segments with negative torsion angle values and all isolated $\beta$-strand segments, i.e., extended segments that are not parallel to other similar motifs. This postprocessing step analyses the global properties of the structure, while the measures of belief/disbelief focus only on the local geometric properties of a motif. Postprocessing drastically reduces the number of incorrectly recognized motifs and consequently increases the quality of the recognition procedure (Rost, Casadia, & Farisellis, 1996).

Table 2 presents a comparison of results of applying the three methods for identifying secondary structure motifs to the experimental electron density map of penicillopepsin.

(1) **2c2c**

| | | | | | | | | | | | | | | |
|---|---|---|---|---|---|---|---|---|---|---|---|---|---|---|
| aa no. | 4 | 5 | 6 | 7 | 8 | 9 | 10 | 11 | 12 | 13 | 14 | 15 | 16 | 17 |
| Actual ss | H | H | H | H | H | H | H | $3_{10}$ | $3_{10}$ | $3_{10}$ | $3_{10}$ | - | - | - |
| MYCIN | H | H | H | H | H | H | H | H | H | H | H | H | H | H |
| $Bayes_1$ | H | H | H | H | H | T | T | H | H | H | H | T | - | - |
| $Bayes_2$ | H | - | H | H | H | H | H | H | H | H | H | H | H | - |

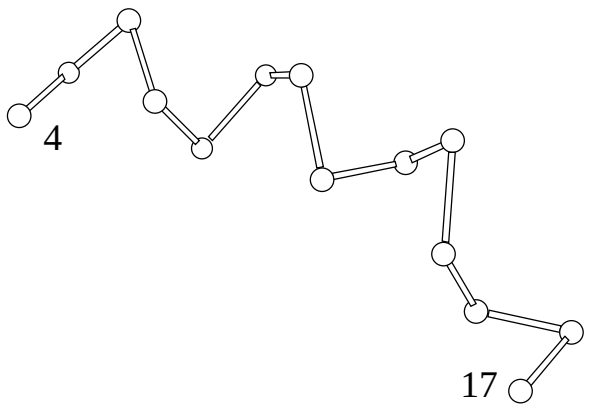

(2) **3app**

| | | | | | | | | | | |
|---|---|---|---|---|---|---|---|---|---|---|
| aa no. | 203 | 204 | 205 | 206 | 207 | 208 | 209 | 210 | 211 | 212 |
| Actual ss | S | S | S | S | S | S | S | S | S | S |
| MYCIN | S | S | S | S | S | S | S | S | S | S |
| $Bayes_1$ | T | - | S | - | - | - | S | S | S | S |
| $Bayes_2$ | S | - | - | - | - | - | S | S | S | S |

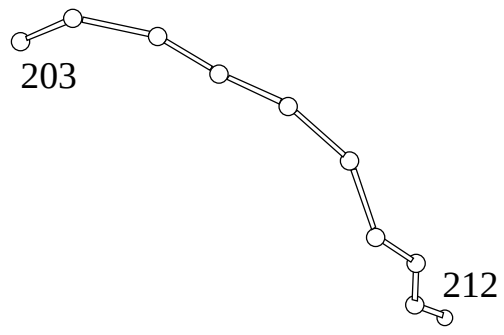

(3) **4rcr**

| | | | | | | | | | | | | | | | |
|---|---|---|---|---|---|---|---|---|---|---|---|---|---|---|---|
| aa no. | 149 | 150 | 151 | 152 | 153 | 154 | 155 | 156 | 157 | 158 | 159 | 160 | 161 | 162 | 163 |
| Actual ss | - | - | - | - | - | H | H | H | H | H | H | H | H | H | H |
| MYCIN | H | H | H | H | H | H | H | H | H | H | H | H | H | H | H |
| $Bayes_1$ | T | S | - | H | T | H | H | H | H | H | H | H | H | H | - |
| $Bayes_2$ | T | - | - | H | T | H | H | H | H | H | H | H | H | H | H |

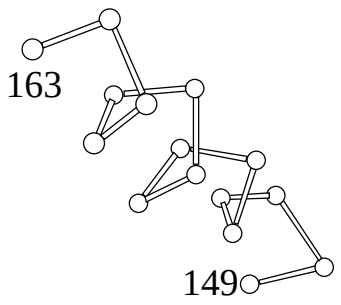

(4) **2c2c**

| | | | | | | | | | |
|---|---|---|---|---|---|---|---|---|---|
| aa no. | 31 | 32 | 33 | 34 | 35 | 36 | 37 | 38 | 39 |
| Actual ss | - | T | T | T | T | T | T | T | - |
| MYCIN | - | H | H | H | H | H | H | - | - |
| $Bayes_1$ | - | T | T | T | T | T | - | T | T |
| $Bayes_2$ | - | T | T | T | T | T | T | - | T |

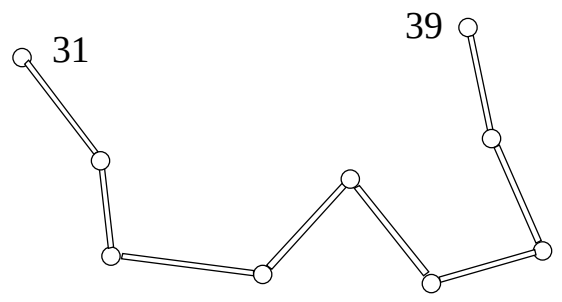

(5) **3app**

| | | | | | | |
|---|---|---|---|---|---|---|
| aa no. | 126 | 127 | 128 | 129 | 130 | 131 |
| Actual ss | - | $3_{10}$ | $3_{10}$ | $3_{10}$ | - | - |
| MYCIN | - | H | H | H | H | - |
| $Bayes_1$ | - | H | H | - | - | - |
| $Bayes_2$ | - | - | - | - | - | - |

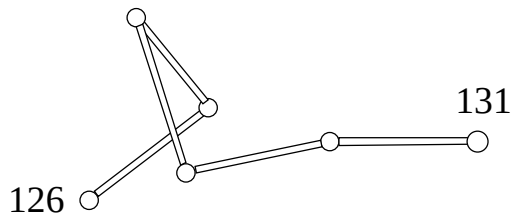

(6) **3app**

| | | | | | | |
|---|---|---|---|---|---|---|
| cp no. | 71 | 199 | 1370 | 506 | 75 | 388 |
| aa no. | 139 | 140 | 141 | 142 | 143 | 144 |
| Actual ss | H | H | H | H | H | H |
| MYCIN | - | H | H | H | H | H |
| $Bayes_1$ | - | T | S | H | - | T |
| $Bayes_2$ | - | H | H | H | - | T |

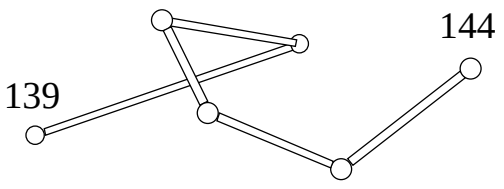

(7) **3app**

| | | | | | | | | | | | | | | | |
|---|---|---|---|---|---|---|---|---|---|---|---|---|---|---|---|
| cp no. | 188 | 482 | 205 | 235 | 905 | 320 | 756 | 581 | 122 | 422 | 1431 | 42 | 630 | 904 | 825 |
| aa no. | 129 | - | 135 | 134 | 133 | 132 | 131 | 132 | - | - | 103 | 102 | - | 85 | 86 |
| Actual ss | H | - | - | - | - | - | - | - | - | - | - | - | - | - | - |
| MYCIN | H | H | H | H | H | H | H | H | H | H | H | H | H | H | H |
| $Bayes_1$ | - | - | - | - | H | T | T | - | - | H | - | H | - | - | - |
| $Bayes_2$ | T | - | H | T | H | T | - | T | - | - | - | H | H | - | H |

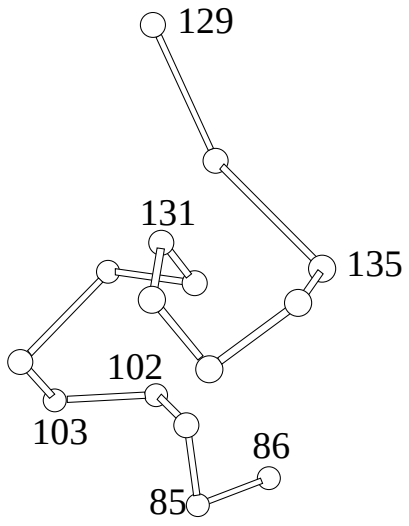

Figure 12: Selected amino acid secondary structure motifs and their identifications ('cp', 'aa', and 'ss' stand for 'critical point', 'amino acid', and 'secondary structure', respectively. 'H', 'T', 'S', and '-' denote the secondary structure classes: 'helix', 'turn', 'sheet', and 'other'.)

| | MYCIN | $\text{Bayes}_1$ | $\text{Bayes}_2$ |
|---|---|---|---|
| $\alpha$-Helices (actual no = 3) | | | |
| No of (correctly) assigned segments | (3) 7 | (0) 0 | (2) 3 |
| % of correctly recognized actual motifs | 91 | 9 | 45 |
| % of correctly identified peaks | 26 | - | 57 |
| $\beta$-Strands (actual no = 15) | | | |
| No of (correctly) assigned segments | (12) 12 | (12) 12 | (9) 9 |
| % of correctly recognized actual motifs | 70 | 41 | 30 |
| % of correctly identified peaks | 73 | 91 | 96 |
| $3_{10}$ Helices (actual no = 2) | | | |
| No of (correctly) assigned segments | (1) 1 | (0) 0 | (0)0 |
| % of correctly recognized actual motifs | 50 | 17 | 0 |
| % of correctly identified peaks | 75 | - | - |
| Turns (actual no = 1) | | | |
| No of (correctly) assigned segments | (0) 0 | (1) 3 | (1) 2 |
| % of correctly recognized actual motifs | 0 | 33 | 27 |
| % of correctly identified peaks | - | 100 | 43 |

Table 2: Results from the application of two Bayesian approaches and a MYCIN-based method to the recognition of secondary structure motifs in minimal spanning trees constructed from a critical point analysis of an experimental electron density map of penicillopepsin at 3 $\mathring{A}$ resolution. Note that the numbers in brackets denote the number of correctly assigned, versus totally assigned, segments (sequences of peaks).

According to Table 2, the MYCIN-based approach appears to have greater success in recognizing helical motifs in experimental maps. Example (6) in Figure 12 depicts one of the three helix motifs that was correctly recognized using the MYCIN-based approach. However, this approach also misidentifies several motifs as helices. Among the four incorrectly identified helices, two four-point segments are actual turns, one four-point segment is characterized by negative torsion angles, and a 15-point sequence of critical points is a succession of three jumps (a jump is a connection occurring between non-adjacent amino acid residues) (See Example (7) in Figure 12). The Bayes modules incorrectly identify a turn in this same region of the electron density map. Jumps are problematic and may seriously hinder the recognition rate, especially in experimental maps blurred by noise and errors.

Table 2 illustrates that the consideration of noisier data in the training set (module $Bayes_2$) leads to an improvement in the number of identified $\alpha$-helices with respect to the first Bayesian module (Example (6) in Figure 12). However, this also leads to a number of incorrectly identified segments. One segment of length two is wrongly identified as helix. It actually corresponds to a jump between non-adjacent amino acid residues; this jump also generates an interpretation error with the MYCIN-based algorithm. The poor accuracy in turn recognition (43 %) is due to this wrongly identified segment.

## 6. Related Research

The interpretation of protein images has been greatly facilitated in recent years by the advent of powerful graphics stations coupled with the implementation of highly efficient computer programs, most notably FRODO (Jones, 1992) and O (Jones et al., 1991). Although these programs were designed specifically for the visual analysis of electron density maps of proteins, they still require a significant amount of expert intervention and interpretation and require considerable time investment. Unlike these systems ORCRIT was designed as a more automated approach to protein model building and interpretation.

The research presented in this paper is a component of an ongoing research project in the area of molecular scene analysis (Fortier et al., 1993; Glasgow et al., 1993). The primary objective of this research is the implementation and application of computational methods for the classification and understanding of complex molecular images. Towards this goal, we have proposed a knowledge representation framework for integrating existing sources of protein knowledge (Glasgow, Fortier, Conklin, Allen, & Leherte, 1995). Representations for reasoning about the visual and spatial characteristics of a molecular scene are captured in this framework using techniques from computational imagery (Glasgow, 1993; Glasgow & Papadias, 1992). This paper extends previous publications in molecular scene analysis by placing the research in an artificial intelligence framework and relating it to work in machine vision. As well, it focuses on the use of uncertain reasoning for secondary structure interpretation in the critical point representation and provides further experimental results supporting our approaches to protein image interpretation.

Two kinds of image improvement procedures are being considered in conjunction with the information stored in a critical point representation. The first one consists of improving phase information at a given resolution. This is a necessary, but difficult, step in a protein structure determination carried out from experimental data. Structural informa-tion retrieved from a topological analysis might be injected into a so-called *direct methods* procedure, which has previously been successfully applied to the structure determination of small molecules at high resolution (Hauptman & Karle, 1953), and more recently to macromolecules as well. However, these methods are presently not applicable to protein images at low- and medium-resolution data, and time-consuming experimental methods are generally used for phase recovery in protein structure elucidation.

The second set of procedures for protein image enhancement involves the construction and interpretation of increasingly higher-resolution maps. This is presently carried out visually by crystallographers who have access to a well-phased medium- or high-resolution map. The highest density regions are fitted with fragments retrieved from a database of chemical templates, eventually allowing for the determination of the protein's 3D structure (Jones et al., 1991). These two protein image reconstruction procedures are interrelated: improved phases lead to a more reliable map in which further motif identification can take place. Under such considerations, secondary structure motifs detected in a low-resolution map are regions of interest to generate medium-resolution information, which would further give access to the the amino acid residue locations.

These procedures give rise to an iterative approach to molecular scene analysis. In an iterative refinement process, if some portion of the image can be interpreted then this information is applied (via an inverse Fourier transform) to adjust the current phases. The modified phases are then used to generate a new image. This approach to scene analysis thus proceeds from an initial coarse (low-resolution) image through progressively more detailed (higher-resolution) images in which further substructures are identifiable[9].

What this implies is that at any particular resolution, it is not necessary that our analysis identify all substructures. The recognition of any parts of the scene can be used to improve phases, giving rise to an overall improvement to the image. The new image can then be further analyzed leading to additional interpretations. This process is iteratively applied (within a heuristic search strategy) until the protein structure is fully determined.

The critical point representation described in this paper is just one component of the knowledge representation scheme for computational imagery. A second component of the scheme involves a spatial logic, which has been used to represent and reason with the concepts and qualitative spatial features associated with a protein molecule (Conklin et al., 1993b; Conklin, Fortier, Glasgow, & Allen, 1996). Associated with the spatial representation is a knowledge discovery technique, called IMEM (Conklin & Glasgow, 1992), based on a theory of conceptual clustering. This system has been used to discover recurrent 3D structural motifs in molecular databases (Conklin, Fortier, & Glasgow, 1993a; Conklin et al., 1996). We anticipate that this and other machine learning/discovery techniques (e.g., (Hunter, 1992; Lapedes, Steeg, & Farber, 1995; Unger et al., 1989)) could be applied to aid in a top-down analysis of novel molecular scenes in order to anticipate and classify structural motifs. This would be complementary to the bottom-up scene analysis provided by the topological approach described in this paper.

Molecular scene analyses can further benefit from research in protein structure prediction. In particular, we are currently investigating formulations derived for the inverse folding problem (Bowie, Luethy, & Eisenberg, 1991). Given an amino acid sequence and a set of

9. The resolution of the image depends both on the number of experimental reflections available as well as on the amount of phase information.

core segments (pieces of secondary structure forming the tightly packed internal protein core), this approach to prediction evaluates each possible alignment (threading) of a known primary sequence of amino acids onto possible core templates. The problem of identifying individual residues in a critical point map at medium to high-resolution can be addressed in a similar manner, i.e., by attempting to thread a sequence onto a protein structure derived from our topological analysis. This problem is significantly simpler than protein structure prediction since it involves threading a sequence onto its own experimentally determined structure, rather than onto templates retrieved from a library of possible models. In the threading approach proposed by Lathrop and Smith (1994), a scoring function is used to derive the statistical preference of a residue for a given environment. Modifications to the current scoring function, to incorporate properties available in the electron density map and critical point graph representations, are being implemented in order to customize this approach to the information available from our topological analysis (Baxter, Steeg, Lathrop, Glasgow, & Fortier, 1996).

## 7. Conclusions

It was reported in this paper that a topological approach can effectively be applied to the segmentation of protein images into their meaningful parts at low- and medium-resolution. Furthermore, it was shown that secondary structure motifs could be identified in medium-resolution images through a geometric analysis of the image representation; the application of geometric rules and probabilities yields a measure of confidence that a given peak is a component of a known secondary structure motif.

Three secondary structure identification modules were applied to the interpretation of ideal and experimental critical point graphs. Two probabilistic Bayesian approaches and a MYCIN-based method all revealed that geometric components such as torsion angles, distances and planar angles are useful in that they differentiate between helices, strands, and turns.

Both the Bayesian and MYCIN-derived approaches were relatively successful in assigning secondary structure identifications to sequences of critical points that are geometrically well resolved. In the case of noisy experimental data, their accuracy decreased. We anticipate that the accuracy could be further improved through the use of larger training sets and training for 3-10 helix and other subclasses. However, we do not expect to achieve a full secondary structure recognition for a protein – rather we expect to interpret good (less noisy) portions of an electron density map and use this information to iteratively improve our image in order to carry out further analyses.

The protein structures used to compose the training and test sets all contain backbone information only. These structures are free of heteroatoms and/or small solvent molecules. The prior knowledge of the chemical composition of a crystallographic cell would certainly help in anticipating problems such as connections between non-adjacent residues through high density peaks. Additional experimental data would permit us to extend the scope of the three approaches described in the present paper.

Modern crystallographic studies remain at the forefront of current efforts to characterize and understand molecular recognition processes. A long-term goal of our research in molecular scene analysis is to assist these studies through a computational methodology for aiding expert crystallographers in the complex imagery processes required to fully interpret the 3D structure of a protein. The topological approach presented here is an important component of this methodology. Further research is required, however, to extend it for the analysis of multi-resolution maps, and to incorporate more domain knowledge into these analyses.

## Acknowledgements

The authors wish to thank Carroll Johnson for providing the ORCRIT program and for his ongoing collaboration with the project, and Marie Fraser for providing access to the experimental electron density map for penicillopepsin. Financial support for this research was provided by the Natural Science and Engineering Research Council (NSERC) of Canada and the Belgian National Council for Scientific Research (FNRS). LL also thanks the FNRS for her "Chargé de Recherches" position.